\documentclass[runningheads]{llncs}
\usepackage{graphicx}
\usepackage{xspace}
\usepackage{wrapfig}
\usepackage{tikz}
\usepackage{comment}
\usepackage{amsmath,amssymb} 
\usepackage{color}
\usepackage{graphicx}
\usepackage{amsmath}
\usepackage{amssymb}
\usepackage{booktabs}
\usepackage{bm}
\usepackage{times}
\usepackage{epsfig}
\usepackage{graphicx}
\usepackage{amsmath}
\usepackage{amssymb}
\usepackage{color,soul}
\usepackage{multirow}
\usepackage{enumitem}
\usepackage[normalem]{ulem}
\usepackage{algorithm}
\usepackage{xcolor}
\usepackage{listings}
\usepackage[font=small]{caption}
\usepackage[font=footnotesize]{subcaption}
\usepackage{stmaryrd}
\usepackage{colortbl}
\usepackage{booktabs} 
\definecolor{mygray}{gray}{.9}

\usepackage[accsupp]{axessibility}  

\usepackage{bm}
\usepackage[accsupp]{axessibility}  

\usepackage[pagebackref=true,breaklinks=true,colorlinks,bookmarks=false,citecolor=blue,linkcolor=blue]{hyperref}

\def\eg{\emph{e.g}.}

\def\etc{\emph{etc}.}

\def\etal{\emph{et al}.}

\begin{document}
\pagestyle{headings}
\mainmatter
\def\ECCVSubNumber{6722}  

\title{DoodleFormer: Creative Sketch Drawing with Transformers} 


\authorrunning{A. K. Bhunia \etal}

\author{Ankan Kumar Bhunia$^1$, Salman Khan$^{1,2}$, Hisham Cholakkal$^{1}$,
Rao Muhammad Anwer$^{1,3}$, Fahad Shahbaz Khan$^{1,4}$, Jorma Laaksonen$^{3}$, Michael Felsberg$^{4}$}
\institute{
  $^1$~Mohamed bin Zayed University of AI, UAE \quad
  $^2$~Australian National University, Australia
  $^3$~Aalto University, Finland
  $^4$~Link{\"o}ping University, Sweden
  \texttt{\small ankan.bhunia@mbzuai.ac.ae} 
  }
\maketitle
  


   \begin{center}
    \includegraphics[width=1\linewidth]{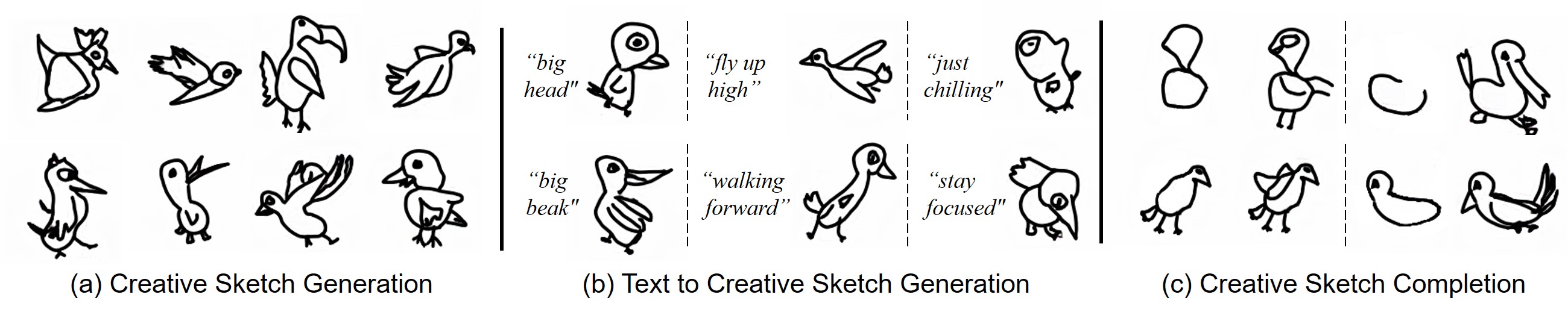}
    \captionof{figure}{Examples of visual creative sketches generated using the proposed DoodleFormer. Here, we show  creative sketches generated based on (a) the random input strokes, (b) text inputs and (c) incomplete sketch images provided by the user. In all three scenarios, the generated sketches are well aligned with the user provided inputs (\eg, the creative sketches generated for the text inputs ``walking  forward"  and ``fly up high" in (b)). Similarly, the diversity in terms of appearance, posture and part size can be observed within the generated creative bird sketches in (a). Furthermore, DoodleFormer accurately completes the missing bird wings, legs and beak in the bottom right example in (c). Additional examples are  available in Fig.~\ref{fig:ganvsformer}, Fig.~\ref{fig:onecol}, Fig.~\ref{fig:qual} and the supplementary.}\label{fig:intro_big}
  \end{center}

\begin{abstract}
Creative sketching or doodling is an expressive  activity, where imaginative and previously unseen depictions of everyday visual objects are drawn.~
Creative sketch image generation is a challenging  vision problem, where the task is to generate diverse, yet realistic creative sketches possessing the unseen composition of the visual-world objects. Here, we propose a novel coarse-to-fine two-stage framework, DoodleFormer, that decomposes the creative sketch generation problem into the creation of coarse sketch composition followed by the incorporation of fine-details in the sketch. We introduce graph-aware transformer encoders that effectively capture global dynamic as well as local static structural relations among different body parts. To ensure diversity of the generated creative sketches, we introduce a probabilistic coarse sketch decoder that explicitly models the variations of each sketch body part to be drawn.
Experiments are performed on two creative sketch datasets: Creative Birds and Creative Creatures. Our qualitative, quantitative and human-based evaluations show that DoodleFormer outperforms the state-of-the-art on both datasets, yielding realistic and diverse creative sketches. On Creative Creatures, DoodleFormer achieves an absolute gain of 25 in
Fr\`echet inception distance (FID) over state-of-the-art. We also demonstrate the effectiveness of DoodleFormer for related applications of text to creative sketch generation, sketch completion and house layout generation.  Code is available at:  \href{https://github.com/ankanbhunia/doodleformer}{https://github.com/ankanbhunia/doodleformer}
\end{abstract}

\section{Introduction}
\label{sec:intro}

Humans have an outstanding ability to easily communicate and express abstract ideas and emotions through sketch drawings. Generally, a sketch comprises several strokes, where each stroke can be considered as a group of points. In automatic sketch image generation, the objective is to generate recognizable sketches that are closely related to the real-world visual concepts. Here, the focus is to learn more canonical and mundane interpretations of everyday objects.  

Different from the standard sketch generation problem discussed above, \textit{creative} sketch generation \cite{ge2020creative} involves drawing more imaginative and previously unseen depictions of everyday visual concepts (see Fig.~\ref{fig:intro_big}(a)). In this problem, creative sketches are generated according to  externally provided random input strokes.  
Example of creative sketch generation includes doodling activity, where diverse, yet recognizable sketch images are generated through unseen composition of everyday visual concepts. 
Automatic generation of creative sketches can largely assist human creative process \eg, inspiring further ideas by providing a possible interpretation of initial sketches by the user. However, such a creative task is more challenging compared to mimicking real-world scenes in to sketch images. This work investigates the problem of creative sketch generation.

\begin{figure*}[t!]
\begin{center}
   \includegraphics[width=\linewidth]{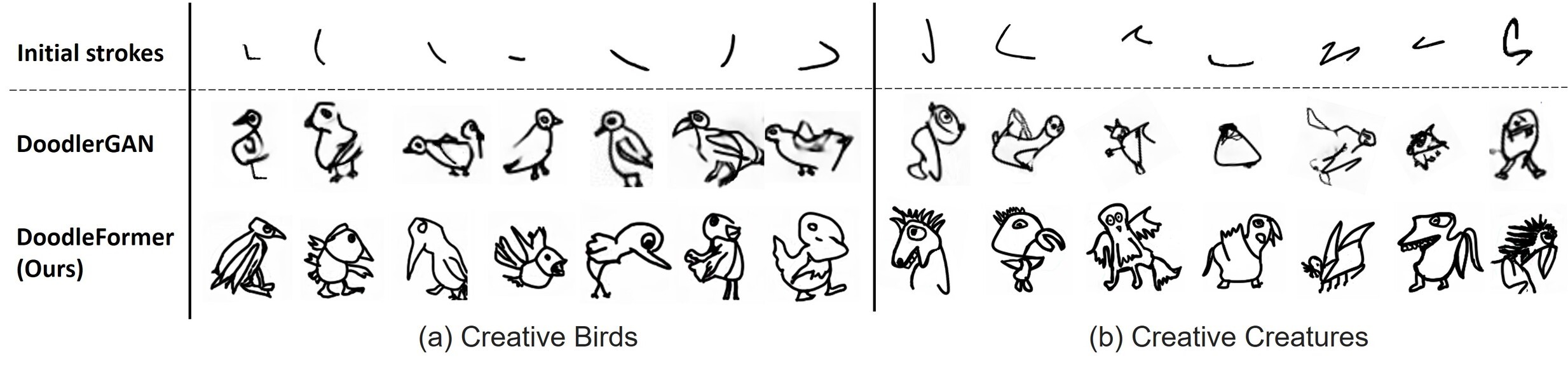}\vspace{-0.7cm}
\end{center}
\caption{A visual comparison of creative sketch images generated by DoodlerGAN \cite{ge2020creative} (top row)  and the proposed DoodleFormer (bottom row) for  the same initial random input strokes. We show examples from both Creative Birds (a) and Creative Creatures (b) datasets. DoodlerGAN suffers from topological artefacts (\eg, more than one head like region in the third bird sketch from the left), disconnected body parts (\eg, the fifth sketch from the left in creatures). Further, the DoodlerGAN generated creative sketches have lesser diversity in terms of size, appearance and posture. The proposed DoodleFormer alleviates the issues of topological artefacts and disconnected body parts, generating creative sketches that are more realistic and diverse.} \vspace{-0.4cm}
\label{fig:ganvsformer}
\end{figure*}

Recently, Ge \etal~\cite{ge2020creative} address the creative sketch image generation problem by proposing a part-based Generative Adversarial Network called DoodlerGAN. It utilizes a part-specific generator to produce each body part of the sketch. The generated body parts are then  sequentially integrated with the externally provided random input, for obtaining final sketch image. Although DoodlerGAN utilizes a part-specific generator for creating each body part of the sketch, it does not comprise an explicit mechanism to ensure that each body part is placed appropriately with respect to the rest of the parts. This leads to topological artifacts and connectivity issues (see Fig.~\ref{fig:ganvsformer}). Further, DoodlerGAN struggles to generate diverse sketch images, which is an especially desired property in creative sketch generation. 

In this work, we argue that the aforementioned problems of topological artefacts, connectivity and diversity issues can be alleviated by imitating the natural \textit{coarse-to-fine} creative sketch drawing process, where the artist first draws the holistic coarse structure of the sketch and then fills the fine-details to generate the final sketch. By first drawing the holistic coarse structure of the sketch aids to appropriately decide the location and the size of each sketch body part to be drawn. To imitate such a coarse-to-fine creative sketch generation process, we look into a two-stage framework where the global as well as local structural relations among different body parts can be first captured at a coarse-level followed by obtaining the fine-level sketch.
The coarse-to-fine 
framework is expected to further improve the diversity of the creative sketch images by explicitly modeling the variations in the location and size of each sketch body part to be drawn.

 

\subsection{Contributions}
We propose a novel two-stage encoder-decoder framework, DoodleFormer, for creative sketch generation. DoodleFormer decomposes the creative sketch generation problem into the construction of holistic coarse sketch composition followed by injecting fine-details to generate final sketch image. To generate realistic sketch images, we introduce graph-aware transformer (GAT) encoders that effectively encode the local structural relations between different sketch body parts by integrating a static adjacency-based graph into the dynamic self-attention block. We further introduce a probabilistic coarse sketch decoder that utilizes Gaussian mixture models (GMMs) to obtain diverse locations of each body part, thereby improving the  diversity of output sketches (see Fig.~\ref{fig:ganvsformer}).

We evaluate the proposed DoodleFormer by conducting extensive
qualitative, quantitative and human-based evaluations on the recently introduced Creative Birds and Creative Creatures datasets. Our DoodleFormer performs favorably against DoodlerGAN on all three evaluations. 
For instance, DoodleFormer sketches were interpreted to be drawn by a human 86\%, having better strokes integration 85\% and being more creative 82\%, over DoodlerGAN in terms of human-based evaluation. 
Further, DoodleFormer outperforms DoodlerGAN with absolute gains of 25  and 23 in terms of Fr\`echet inception distance (FID) on Creative Creatures and Creative Birds, respectively. In addition to sketch generation based on externally provided random initial strokes, we validate the effectiveness of DoodleFormer to generate creative sketches based on text inputs, incomplete sketch images provided by user as well as generating complete house layouts given coarse-level bubble diagrams. DoodleFormer achieves impressive performance for text to sketch generation, sketch completion (see Fig.~\ref{fig:intro_big} (b) and (c)) as well as house layout generation (see Fig.~\ref{fig:hsgan}).

\section{Related Work}
The problem of sketch generation ~\cite{ha2017neural,hinton2006inferring,li2017free,cao2019ai,zheng2018strokenet,chen2017sketch} has been studied extensively in literature. These methods generally aim to mimic the visual world by capturing its important aspects in the generated sketches.  SketchRNN~\cite{ha2017neural}, utilizes sequence-to-sequence Variational Autoencoder (VAE) for conditional and unconditional generation of vector sketches.  
Cao \etal~\cite{cao2019ai} propose a generative model that generates multi-class sketches. Moreover, alternative strategies such as differentiable rendering~\cite{zheng2018strokenet}, 
attention-based architectures \cite{ribeiro2020sketchformer} and reinforcement learning \cite{balasubramanian2019teaching,zhou2018learning,ganin2018synthesizing} 
have been investigated for sketch generation. The work of \cite{chen2017sketch} incorporates a convolutional encoder to capture the spatial layout of sketches, whereas \cite{liu2019sketchgan,su2020sketchhealer,qi2021sketchlattice} aim at completing the missing parts of  sketches. The work of \cite{lin2020sketch} targets recovering the masked parts of points in sketches. 
A few works \cite{ribeiro2020sketchformer,lin2020sketch} have also studied related tasks of sketch classification and retrieval.

Different from the aforementioned standard sketch generation task, creative sketch generation has been recently explored \cite{ge2020creative}. This task focuses on drawing more imaginative and previously unseen depictions of common visual concepts rather than generating  canonical and mundane interpretations of visual objects. To this end, DoodlerGAN  \cite{ge2020creative} introduces a part-based Generative Adversarial Network built on StyleGAN2 \cite{Karras2019stylegan2} to sequentially produce each body part of the creative sketch image.  Here, the part-based GAN model  needs to be trained separately for individual body parts (eye, head, beak, \etc) using part annotations. However, such a separate model for each body part results in a large computational overhead. 
During inference, these individual part-based GAN models are sequentially used to  generate their respective body parts within the creative sketch. While generating recognizable creative sketches, DoodlerGANs struggles with topological artifacts, connectivity and diversity issues. In this work, we set out to overcome these issues to generate diverse, yet realistic creative sketches.

\section{Our Approach}
\textbf{Motivation:}
To motivate our framework, we first distinguish two desirable properties to be considered when designing an approach for creative sketch generation.

\begin{figure*}[t!]
\begin{center}
   \includegraphics[width=1\textwidth]{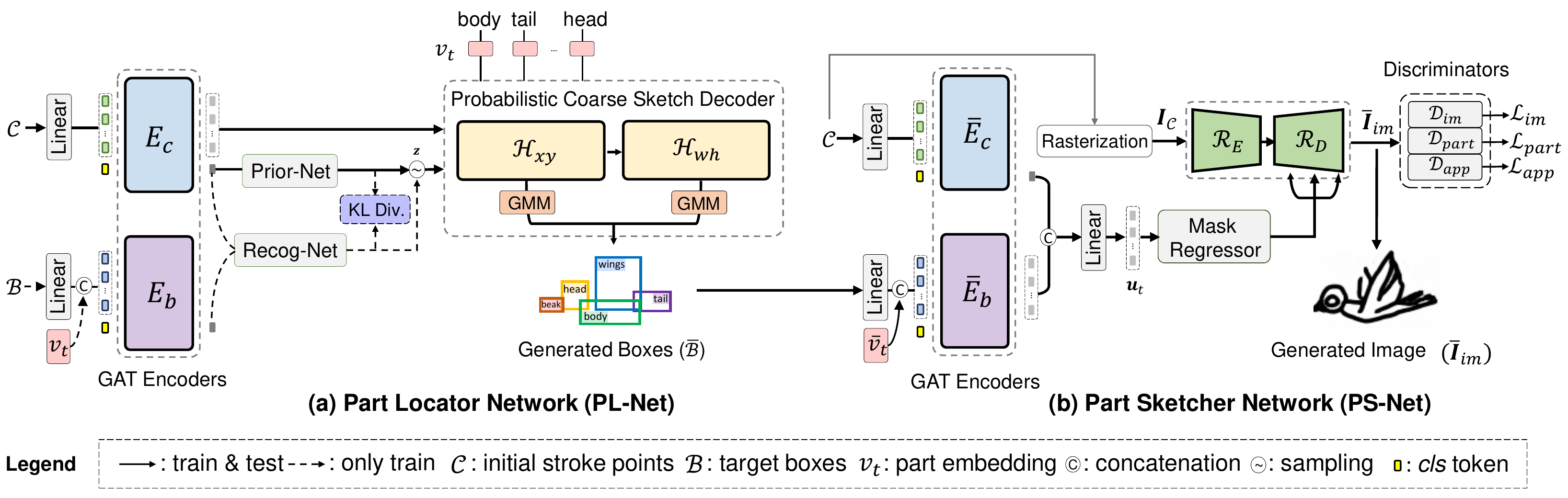}\vspace{-0.5cm}
\end{center}
\caption{The proposed DoodleFormer comprises two stages: Part Locator (PL-Net) and Part Sketcher (PS-Net). (a) The first stage, PL-Net, takes the initial stroke points $\mathcal{C}$ as the conditional input and learns to return the bounding boxes corresponding to each body part (coarse structure of the sketch) to be drawn. PL-Net contains two graph-aware transformer (GAT) encoders  ($E_b$,  $E_c$) and a probablistic coarse sketch decoder utilizing GMM modelling for the coarse box prediction. Within the decoder,  the bounding box parameters are predicted by the location-predictor ($\mathcal{H}_{xy}$) and size-predictor ($\mathcal{H}_{wh}$) modules. 
(b) The second stage, PS-Net, then takes the predicted box locations along with $\mathcal{C}$ as inputs  and generates the final sketch image $\bar{\bm{I}}_{im}$.
Following the design of $E_b$ and  $E_c$, PS-Net also comprises GAT block-based encoders ($\bar E_b$, $\bar E_c$). Further, PS-Net contains a convolutional encoder-decoder network ($\mathcal{R}_E$, $\mathcal{R}_D$) and a mask regressor  to generate rasterized high quality sketch image $\bar{\bm{I}}_{im}$. 
} \vspace{-0.4cm}
\label{fig:method}
\end{figure*}

\noindent\textbf{\textit{Holistic Sketch Part Composition:}} As discussed earlier, DoodlerGAN employs a part-specific generator to produce each body part of the sketch. However, it does not utilize any explicit mechanism to ensure that the generated part is placed in an appropriate location relative to other parts, thereby suffering from topological artifacts and connectivity issues (see Fig.~\ref{fig:ganvsformer}). Here, we argue that explicitly capturing the holistic arrangement of the sketch parts is desired to generate realistic sketch images that avoid topological artifacts and connectivity issues. \\
\noindent\textbf{\textit{Fine-level Diverse Sketch Generation}:} Creative sketches exhibit a large diversity in appearance, posing a major challenge when generating diverse, yet realistic fine-detailed sketch images. Existing work of DoodlerGAN  struggles to generate diverse sketch images since it typically ignores the noise input in the sketch generation process \cite{ramasinghe2021train}. Although DoodlerGAN  attempts to partially address this issue by introducing heuristics in the form of randomly translating the input partial sketch, the diversity of generated sketch images is still far from satisfactory (see Fig.~\ref{fig:ganvsformer}). Instead, we argue that having an explicit probabilistic modeling within the framework is expected to further improve the diversity of the generated sketch images. \\
\textbf{Overall Framework:}
The proposed two-stage DoodleFormer framework combines the two aforementioned desired properties by decomposing the creative sketch generation problem to first capture the holistic coarse sketch composition and then injecting fine-details to generate the final sketch. The overall architecture of the proposed two-stage DoodleFormer is shown in Fig.~\ref{fig:method}. DoodleFormer comprises two stages: \emph{Part Locator} (PL-Net) and \emph{Part Sketcher} (PS-Net). 
The first stage, Part Locator (PL-Net), learns to explicitly capture the holistic arrangement of the sketch parts conditioned on the externally provided random initial stroke points $\mathcal{C}$ represented in a vector form. 
PL-Net comprises graph-aware transformer (GAT) block-based encoders to capture structural relationship between different regions within a sketch. To the best of our knowledge, we are the first to introduce a GAT block-based transformer encoder for the problem of creative sketch image generation. 
Instead of directly predicting the box parameters as deterministic points from the transformer decoder, we further introduce probabilistic coarse sketch decoders that utilize GMM modelling for box prediction.
This enables our DoodleFormer to achieve diverse, yet plausible coarse structure (bounding boxes) for sketch generation.
The second stage, Part Sketcher (PS-Net), creates the final sketch image with appropriate line segments based on the coarse structure obtained from PL-Net.
PS-Net also comprises GAT block-based encoders, as in PL-Net, along with a convolutional encoder-decoder network to generate  the final rasterized sketch image.  

\begin{figure*}[t!]
\begin{center}
   \includegraphics[width=1\textwidth]{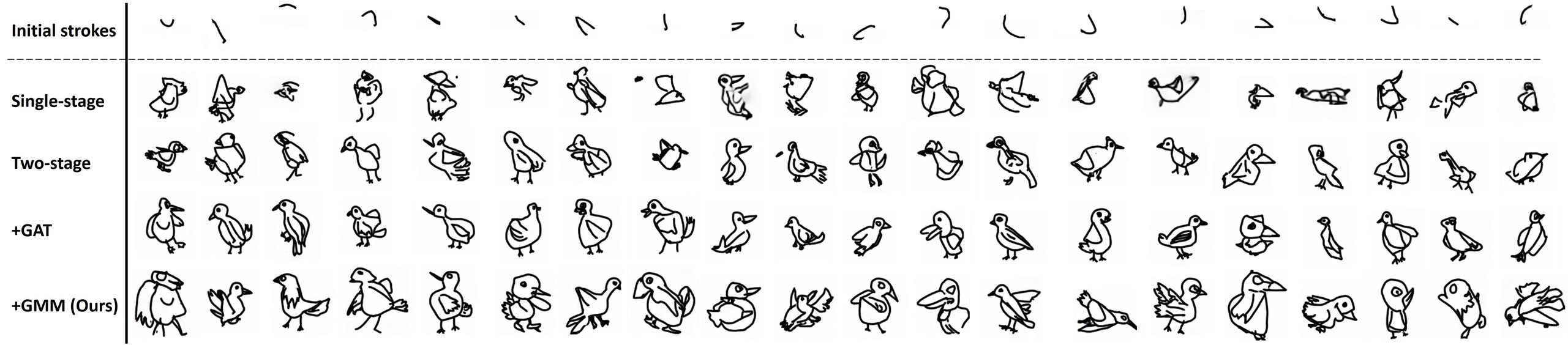}\vspace{-0.5cm}
\end{center}
\caption{A visual comparison in terms of progressively integrating one contribution at the time, from top to bottom, for common initial strokes. Compared to the single-stage baseline (first row), the two-stage framework (without the GAT block and probabilistic modeling in the decoder) generates sketch in a coarse-to-fine manner. As a result, the two-stage framework (second row) produces a more complete sketch where each body part is placed at an appropriate location relative to other parts. The introduction of GAT block (third row) in the encoders of the two-stage framework improves the realism of the generated sketches by capturing the structural relationship between different parts (\eg, the tenth image from the left, where there is a discontinuity between the beak and the head of the bird). Further, the introduction of probabilistic modelling in the decoder of the two-stage framework (last row), improves the diversity (\eg, appearance, size, orientation and posture) of generated sketch images. Our final two-stage framework (last row) produces realistic and diverse sketch images.   
} \vspace{-0.4cm}
\label{fig:onecol}
\end{figure*}

Our carefully designed two-stage DoodleFormer architecture possesses both desired properties (holistic sketch part composition as well as fine-level diverse sketch generation) and creates diverse, yet realistic sketch images in a coarse-to-fine manner (see Fig.~\ref{fig:onecol}). 
Next, we describe in detail PL-Net (Sec.~\ref{Sec:PL-Net}) and PS-Net (Sec.~\ref{Sec:PS-Net}). 

\subsection{Part Locator Network (PL-Net)}
\label{Sec:PL-Net}

As discussed above, PL-Net takes the initial stroke points $\mathcal{C}$ as the conditional input, and learns to return a coarse structure capturing the holistic part composition of the desired sketch. The \emph{encoders} in PL-Net contain graph-aware transformer (GAT) blocks  to encode the structural relationship between different parts (holistic sketch part composition), leading to realistic sketch image generation. The \emph{decoder} in PL-Net utilizes GMM modeling for box prediction, enabling the generations of diverse sketch images. 

\subsubsection{Graph-aware Transformer Block-based Encoder} 
PL-Net consists of two graph-aware transformer (GAT) block-based encoders $E_b$ and $E_c$, which are used to obtain contextualized  representation of the coarse (holistic) structure $\mathcal{B}$ and the conditional input $\mathcal{C}$, respectively. 

\begin{wrapfigure}{r}{0.5\textwidth} 
\vspace{-30pt}
  \begin{center}
    \includegraphics[width=0.5\textwidth]{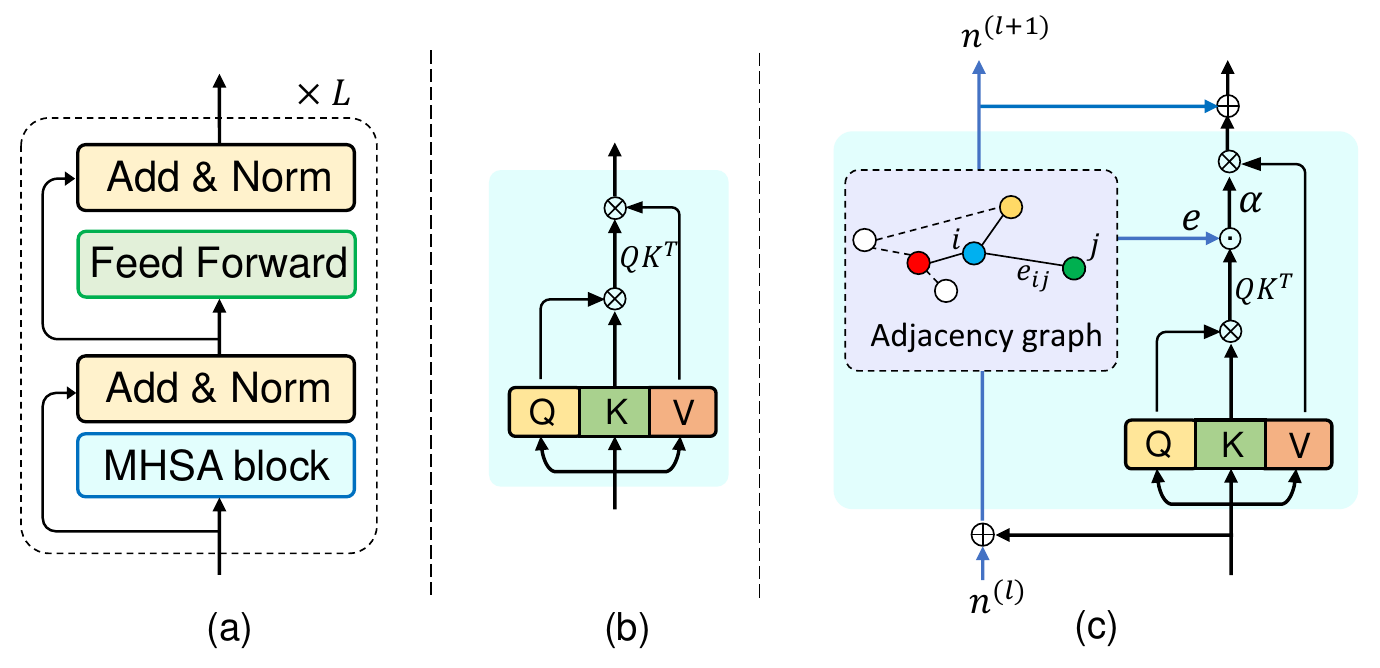}
    \caption{Our proposed graph-aware transformer (GAT) block (c) replaces the standard self-attention (b) in the conventional transformer encoder layer (a). Our GAT block injects the graph structure into self-attention by learning to re-weight the attention matrix based on the pair-wise relations between the graph nodes. In this way, the proposed GAT block combines the local connectivity patterns from the learned adjacency graph with the dynamic attention from the self-attention block.}
    
    
    \label{fig:gatr}
  \end{center}
  \vspace{-20pt}
  \vspace{1pt}
\end{wrapfigure}
 To  encode the  identity $t$ of each body part present in a sketch, we define $\bm{v}_{t} \in \mathbb{R}^{d} $ as a learned part embedding. We concatenate $\bm{v}_{t}$  with a feature representation obtained from  $\bm{b}_{t} \in \mathcal{B}$
 (box location and size information ($x_t, y_t, w_t, h_t$) of each body part). This concatenated feature is then used as an input to the encoder  $E_b$. 
 The conditional input strokes $\mathcal{C}$ are passed through a linear layer before being input  to the encoder $E_c$. 
We add  special $cls$ tokens  \cite{devlin2018bert} at the beginning of input sequences to the encoders ($E_b$ and $E_c$). The output of this token is considered as the contextualized representation of the whole sequence. Further, we use fixed positional encodings to the input of each attention layer to retain information regarding the sequence order. 
Next, we introduce our GAT block used in both encoders ($E_b, E_c$) to encode the holistic structural composition of (sketch) body parts.

\noindent\textbf{Graph-aware Transformer (GAT) Block:}\label{sec:gat}
The structure of our GAT block is shown in Fig.~\ref{fig:gatr}(c).  Each GAT block consists of a graph-aware multi-headed self-attention (MHSA) module followed by a feed-forward network \cite{khan2021transformers}. Given the queries $\bm{Q}$, keys $\bm{K}$, and values $\bm{V}$ the standard self-attention module \cite{vaswani2017attention} computes the attention according to the following equation  (also shown in Fig.~\ref{fig:gatr}(b)),\vspace{-0.2em}
\begin{equation}
\label{eq_13}
\bm{\alpha} =  \textrm{softmax}\left(\frac{\bm{Q}\bm{K}^T}{\sqrt{d}}\right).
\end{equation}

While the standard self-attention module is effective towards learning highly contextualized feature representation, it does not explicitly emphasize on the local structural relation. However, creative sketches are structured inputs with  definite connectivity patterns between sketch parts. To model this structure, we propose to encode an adjacency based graph implemented with spectral graph convolution  \cite{kipf2016semi}. Our proposed GAT block combines the definite connectivity patterns from the learned adjacency graph with the dynamic attention from self-attention block. 
Let us consider a graph where each node $i$ is associated with a structure-aware representation $\bm{n}_i$ and corresponding neighbour set $\mathcal{N}_r(i)$. To represent the neighbor set $\mathcal{N}_r(i)$ for each node $i$, we define an adjacency matrix $\bm{A}$ where each entry represents whether two nodes $i$ and $j$ are adjacent. The edge weight $e_{ij}$ between two adjacent nodes $i$ and $j$ is given by,  
\begin{equation}
\label{eq_7}
e_{ij} = \bm{W}_b^{T} \textrm{ReLU}\left(\bm{W}_a \left[\bm{n}_i , \bm{n}_j\right]\right) \; \forall j \in \mathcal{N}_r(i),
\end{equation}
where $\bm{W}_a$ and $\bm{W}_b$ are learned parameters and $[\cdot,\cdot]$ is a concatenation operator. We set $e_{ij} = 0$ $\forall j \notin \mathcal{N}_r(i)$. For each GAT block $l$, the spectral graph convolution operation is, 
\begin{equation}
\label{eq_9}
\bm{n}_i^{(l+1)} = \textrm{ReLU} \left(\bm{n}_i^{(l)} + \sum_{j \in \mathcal{N}_{r}(i)}^{}e_{ij}\bm{W}_{c}\bm{n}_j^{(l)}\right),
\end{equation}
{where $\bm{W}_c$ is a learned matrix.}
Our main intuition is that the adjacency matrix representing the neighbourhood graph structure is static which is computed over the connected components in the graph and predetermined for each input, it is also symmetric and generally sparse. In contrast, attention learned from the self-attention layer is dynamic, can be dense and also non-symmetric. We propose to combine these two complementary representations through the following equation where we calculate the attention weight $\alpha_{ij}$ for nodes $j \in \mathcal{N}_r(i)$ as follows,
\begin{equation}
\label{eq_8}
\alpha_{ij} = \frac{e_{ij}\exp (\varphi_{ij})}{\sum_{j \in \mathcal{N}_{r}(i)}^{} e_{ij}\exp (\varphi_{ij})},\; \text{s.t. }  \; \varphi_{ij} \in \frac{\bm{Q}\bm{K}^T}{\sqrt{d}}, 
\end{equation}
where $\varphi_{ij}$ is an element of the standard attention matrix.

The special token ($cls$)  output from $E_c$ is then utilized as an input to a  Prior-Net for  approximating the conditional prior latent distribution. Similarly, the $cls$ token outputs of both $E_b$ and $E_c$  are provided as input to a Recog-Net  for approximating the variational latent distribution.
Both the  Prior-Net  and the Recog-Net  are parameterized by multi-layer perceptrons (MLPs) to approximate prior and variational latent normal distributions. 
During training, we sample the latent variable $\bm{z}$ from the variational distribution and provide it as input to the probabilistic coarse sketch decoder. 

\subsubsection{Probabilistic Coarse Sketch Decoder}
The probabilistic coarse sketch decoder within our PL-Net utilizes probabilistic modelling to generate diverse coarse structure. The decoder comprises two modules: a location-predictor $\mathcal{H}_{xy}$ and a size-predictor $\mathcal{H}_{wh}$. Here, the location-predictor $\mathcal{H}_{xy}$ estimates the center coordinates ($x_t,y_t$) of bounding boxes around  body parts, while the size-predictor $\mathcal{H}_{wh}$ predicts their width and height ($w_t, h_t$). 
Both these modules consist of multi-headed self- and encoder-decoder attention mechanisms \cite{vaswani2017attention}.  The encoder-decoder attention obtains the key and value vectors from the output of the encoder $E_c$. This allows every position in the decoder to attend to all positions in the conditional input sequence. The part embedding $\bm{v}_{t}$ from the encoder is used as a query positional encoding to each attention layer of the decoder. Over multiple consecutive decoding layers, the decoder modules produce respective output features $\bm{f}_t^{xy} \in \mathbb{R}^{d}$ and $\bm{f}_t^{wh}\in \mathbb{R}^{d}$ that lead to the distribution parameters of bounding boxes being associated with each body part, representing the coarse structure of the final sketch to be generated. 

To enhance the diversity of generated sketch images, we model the box predictions from each decoder module by Gaussian Mixture Models (GMMs) \cite{bishop1994mixture,graves2013generating}.  
Different from the conventional box prediction \cite{carion2020end,zhu2020deformable} that directly maps the {decoder} output features as deterministic box parameters, our GMM-based box prediction is modeled with $M$ normal distributions $\mathcal{N}\left(\cdot \right)$ where each distribution is parameterized by $\theta_{k}$  and a mixture weight $\pi_{k}$,
\begin{equation}
\begin{split}
\label{eq_4}
     p(\bm{b}_t|\mathcal{C}, \bm{z}) = \sum_{k=1}^{M}\pi_{k,t}\mathcal{N}\left(\bm{b}_t; \theta_{k,t}\right),   \textrm{for} \sum_{k=1}^{M}\pi_{k,t} = 1.
\end{split}
\end{equation} 
The GMM parameters can be  obtained by minimizing the negative log-likehood  for all $P$ body parts in a sketch, 
\begin{equation}
\label{eq_10}
     \mathcal{L}_{b} = -\frac{1}{P}\sum_{t=1}^{P}\log\left( \sum_{k=1}^{M}\pi_{k,t}\mathcal{N}(\bm{b}_t; \theta_{k,t})\right) .
\end{equation} 
 Here, we  simplify the quadvariate distribution of GMMs in Eq.~\ref{eq_10} by decomposing it into two bivariate distributions as $p(\bm{b}_t|\mathcal{C},\bm{z})=p(x_t,y_t|\mathcal{C}, \bm{z})p(w_t,h_t|x_t, y_t,\mathcal{C}, \bm{z})$. 
The  parameters of these bivariate GMMs are  obtained by employing linear layers and  appropriate normalization on the outputs $f_t^{xy}$,  $f_t^{wh}$  of  $\mathcal{H}_{xy}$ and  $\mathcal{H}_{wh}$, respectively.  
In addition to GMM parameters, these linear layers also estimate the presence of a body part using an indicator variable, which  is trained with a binary cross entropy loss $\mathcal{L}_{c}$. 

\noindent\textbf{PL-Net Loss function ($\mathcal{L}_{PL}$):} The overall loss function $\mathcal{L}_{PL}$ to train the PL-Net is the weighted sum of the reconstruction loss $\mathcal{L}_{rec}$, and the KL divergence loss $\mathcal{L}_{KL}$ ,
\begin{equation}
\label{eq_11}
     \mathcal{L}_{PL} = \mathcal{L}_{rec} + \lambda_{KL}\mathcal{L}_{KL}.
\end{equation} 
{Here, the reconstruction loss term is $\mathcal{L}_{rec} = \mathcal{L}_b + \mathcal{L}_c$.
The KL divergence loss term $\mathcal{L}_{KL}$ regularizes the variational distribution \cite{sohn2015learning}  from the Recog-Net to be closer to the prior distribution from the conditional Prior-Net, whereas  $\lambda_{KL}$ is a scalar loss weight.} 

Our carefully designed PL-Net architecture, presented above, provides a coarse structure of the sketch that is used to generate a diverse, yet realistic final sketch image in the second stage (PS-Net) of the proposed two-stage DoodleFormer framework. Next, we present the PS-Net that takes the coarse structure of the sketch along with initial partial sketch $\mathcal{C}$  as inputs and generates the final sketch image.  



\subsection{Part Sketcher Network (PS-Net)}
\label{Sec:PS-Net}
Our PS-Net comprises two graph-aware transformer (GAT) block-based encoders $\Bar{E}_b$ and $\Bar{E}_c$, following the design of encoders ${E}_b$ and ${E}_c$ in the PL-Net. 
Here, the encoder $\Bar{E}_b$ produces a contextualized  feature representation of bounding box $\bm{b}_t$ associated with each body part. Similarly, the encoder $\Bar{E}_c$ outputs a  contextualized feature representation of initial stroke points $\mathcal{C}$.  Both these contextualized feature representations from $\Bar{E}_b$ and $\Bar{E}_c$ are  then concatenated and  passed  through a linear layer to obtain $\bm{u}_t$.  



The initial stroke points $\mathcal{C}$ is converted to its raster form $\bm{I}_\mathcal{C}$ and passed through a {convolutional} encoder  $\mathcal{R}_E$ that outputs a spatial representation $\bm{g} = \mathcal{R}_E(\bm{I}_\mathcal{C})$. 
Consequently, $\bm{g}$ and $\{\bm{u}_t\}_{t=1}^{P}$ are  provided as input to a {convolutional} decoder  $\mathcal{R}_D$ for generating the final sketch image $\bar{\bm{I}}_{im}$,
\begin{equation}
\label{eq_5}
\bar{\bm{I}}_{im} = \mathcal{R}_D\left( \bm{g}, \{\bm{u}_t\}_{t=1}^{P}\right).
\end{equation} 
The decoder network $\mathcal{R}_D$ utilizes the ResNet \cite{he2016deep} architecture as a backbone. To introduce diversity in the generated images, a zero-mean unit-variance multivariate random noise is added with $\bm{g}$ before passing it to the decoder network. For fine-grained shape prediction, we utilize a mask regressor  \cite{sun2019image,sun2020learning} having up-sampling convolutions, followed by sigmoid transformation to generate an auxiliary mask 
for each bounding box. The predicted masks are resized to the sizes of corresponding bounding boxes, which are then used to compute the  instance-specific and structure-aware affine transformation parameters in the normalization layer {of the decoder $\mathcal{R}_D$}. 

The training of PS-Net follows the standard GAN formulation where the PS-Net generator $\mathcal{G}$ is followed by additional discriminator networks $\mathcal{D}_{im}$, $\mathcal{D}_{part}$, and $\mathcal{D}_{app}$ to obtain image-level ($\mathcal{L}_{im}$),  part-level ($\mathcal{L}_{part}$), and appearance ($\mathcal{L}_{app}$) adversarial losses \cite{he2021context,sun2019image},  respectively. The loss function is then given by, 
\begin{equation}
\begin{split}
\label{eq_6}
\mathcal{L}_{PS} = & \mathcal{L}_{im} + \lambda_p \mathcal{L}_{part} + \lambda_a \mathcal{L}_{app},\\
\end{split}
\end{equation}
where $\lambda_p$ and $\lambda_a$ are the loss weight hyper-parameters. 

The introduction of the GAT block in the PL-Net and PS-Net encoders contributes towards the generation of realistic sketch images, whereas the effective utilization of probabilistic modelling in the PL-Net decoder leads to improved diversity. In summary, our two-stage DoodleFormer generates diverse, yet realistic sketch images (see Fig.~\ref{fig:onecol}).

\section{Experiments}
\noindent\textbf{Datasets:} We perform extensive experiments on the recently introduced Creative Birds and Creative Creatures datasets \cite{ge2020creative}. The Creative Birds has 8067 sketches of birds, whereas the Creative Creatures contains 9097 sketches of various creatures. In both datasets, all sketches come with part annotations. Both datasets also contain free-form natural language phrase as a text description for each sketch. \\
\noindent\textbf{Implementation Details:} As discussed, both PL-Net and PS-Net utilize graph-aware transformer (GAT) block-based encoders. In each GAT-block based encoder, we define an adjacency matrix $\bm{A}$ based on the connectivity patterns of the adjacency graph. Every pair of overlapping bounding boxes on a coarse structure is connected in the adjacency graph. Similarly, for initial strokes, the corresponding adjacency graph connects adjacent points on each single stroke. Each of these encoders consist of $L{=} 6$ graph-aware transformer blocks.  Here, each block comprises multi-headed attention having 8 heads. In the probabilistic coarse sketch decoder,   the location-predictor and size-predictor utilize  3 self- and encoder-decoder attention layers. Further,  we set the embedding size $d{=}512$. We augment the vector sketch images by applying small affine transformations and these  vector sketches are converted to raster images of size $128 \times 128$. Our DoodleFormer is trained as follows. In the first stage, for training PL-Net, we initially obtain the bounding boxes for all body parts in a sketch using the part annotations. The bounding boxes are normalized to values between 0 and 1. In the second stage, we train PS-Net using the raster sketch images and their corresponding ground-truth  boxes. In both stages, the initial stroke points are provided in a vector form as a conditional input. 
In all experiments, we use a batch size of 32. The learning rate is set to $1e^{-4}$ and the loss weights $\lambda_{KL}$, $\lambda_{p}$, $\lambda_{a}$ are set to 1, 10 and 10.

\begin{table}[t!]
\begin{center}
\caption{
\textbf{Comparison of DoodleFormer with DoodlerGAN~\cite{ge2020creative}, StyleGAN2~\cite{Karras2019stylegan2} and SketchRNN~\cite{ha2017neural}} in terms of Fr\`echet inception distance (FID), generation diversity (GD), characteristic score (CS) and semantic diversity score (SDS). Our DoodleFormer performs favorably against existing methods on both datasets. 
}\vspace{0.2cm}
\label{tab:main_table}
\setlength{\tabcolsep}{6pt}
\scalebox{0.7}{
\begin{tabular}{l|c|c|c|c|c|c|c}
\toprule[0.4mm]

\rowcolor{mygray} 
\cellcolor{mygray} &
  \multicolumn{3}{c|}{\cellcolor{mygray}Creative Birds} &
  \multicolumn{4}{c}{\cellcolor{mygray}Creative Creatures} \\ 
\rowcolor{mygray} 
\multirow{-2}{*}{\cellcolor{mygray}Methods} &
  FID($\downarrow$) & GD($\uparrow$) & CS($\uparrow$) & FID($\downarrow$) & GD($\uparrow$) & CS($\uparrow$) & SDS($\uparrow$)\\ \midrule
  
Training Data  &- &19.40 &0.45 &- &18.06 &0.60 &1.91 \\   \midrule
SketchRNN~\cite{ha2017neural}       &82.17 &17.29 &0.18 &54.12 &16.11 &0.48 &1.34 \\
StyleGAN2~\cite{Karras2019stylegan2}     &130.93 &14.45 &0.12 &56.81 &13.96 &0.37 &1.17 \\
DoodlerGAN~\cite{ge2020creative}      &39.95 &16.33 &\textbf{0.69} &43.94 &14.57 &0.55 &1.45 \\\midrule
\textbf{DoodleFormer (Ours)}          &\textbf{16.45} &\textbf{18.33} &0.55 &\textbf{18.71} &\textbf{16.89} &\textbf{0.56} &\textbf{1.78} \\  \bottomrule[0.4mm]
\end{tabular}
}
\end{center} \vspace{-0.3cm}
\end{table}

\begin{figure*}[t!]
\begin{center}
   \includegraphics[width=\linewidth]{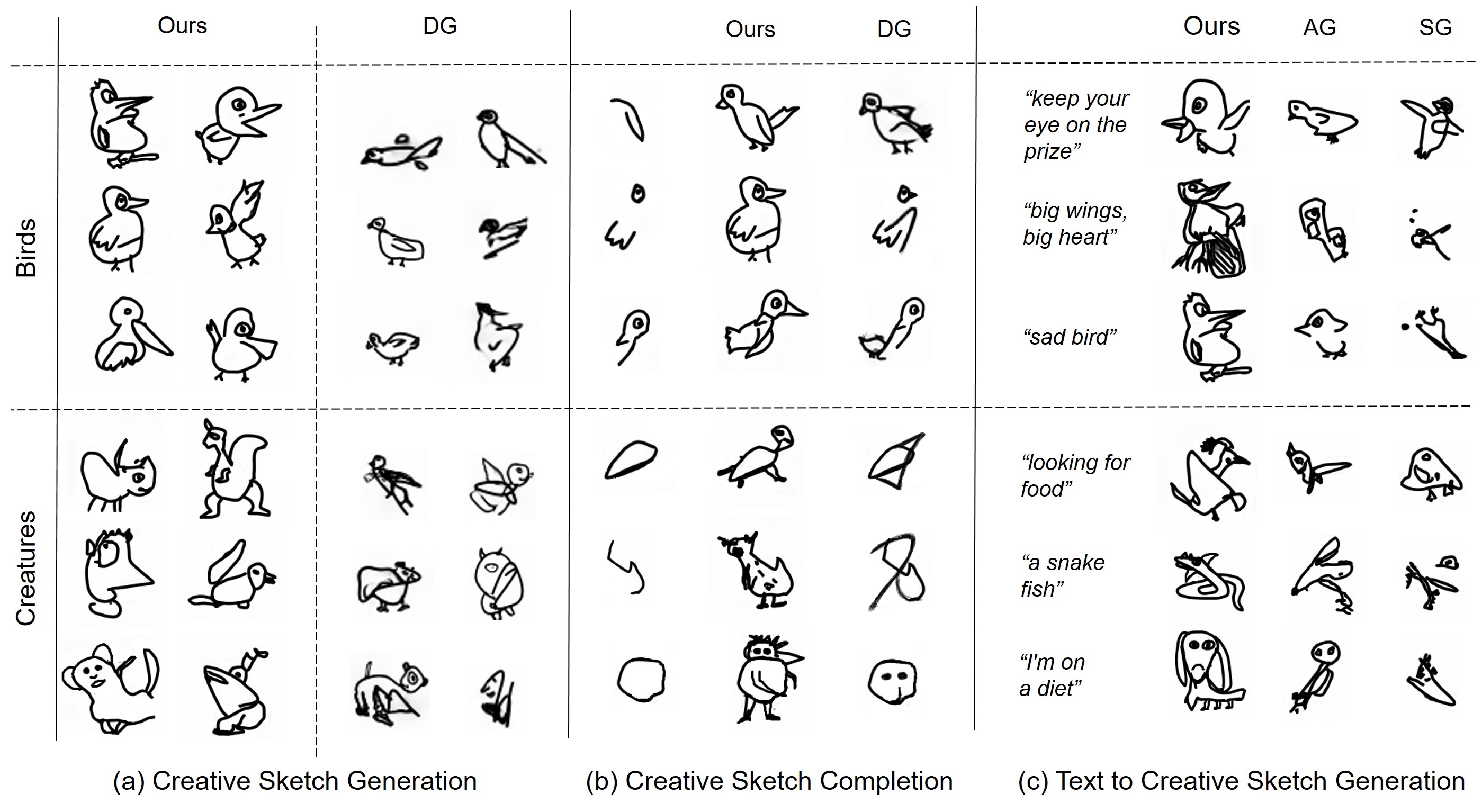}\vspace{-0.5cm}
\end{center}
\caption{Applications of DoodleFormer. (a) Creative sketch generation based on random input strokes. (b) Creative sketch completion: Here, DoodleFormer accurately completes missing parts (\eg, beak, head and body of bird is well connected in second row and column in (b)), compared to DoodlerGAN (DG). (c) Text to creative sketch generation: We compare DoodleFormer with AttnGAN~\cite{xu2018attngan} (AG), StackGAN~\cite{zhang2017stackgan} (SG). DoodleFormer produces sketches that are well aligned with user provided input texts. Best viewed zoomed in.
}
\vspace{-0.3cm}
\label{fig:qual}
\end{figure*}

\subsection{Quantitative and Qualitative Comparisons}
We first present a comparison (Tab.~\ref{tab:main_table}) of our  DoodleFormer  with state-of-the-art approaches~\cite{ha2017neural,viazovetskyi2020stylegan2,ge2020creative} on both Creative Birds and Creative Creatures. For a fair comparison, we evaluate all methods using two widely used metrics, namely Fr\`echet inception distance (FID) ~\cite{heusel2017gans} and generation diversity (GD) ~\cite{cao2019ai}, as in DoodlerGAN~\cite{ge2020creative}. Here,  the FID and GD scores are computed using an Inception model trained on the QuickDraw3.8M dataset ~\cite{xu2020deep}, that embeds the images onto a feature space ~\cite{ge2020creative}.  Tab.~\ref{tab:main_table} shows  that DoodleFormer outperforms existing methods in terms of both  FID and GD scores, on the two datasets. The higher GD score indicates the ability of DoodleFormer to generate diverse sketch images, whereas the  lower  FID score indicates the superior quality of its generated creative sketches.

Furthermore, similar to  DoodlerGAN \cite{ge2020creative}, we  use two additional metrics: characteristic score (CS) and semantic diversity score (SDS). The CS metric evaluates how often a generated sketch is classified to be a bird (for Creative Birds) or creature (for Creative Creatures) by the Inception model trained on the QuickDraw3.8M dataset.  The SDS metric measures the diversity of the sketches in terms of the different creature categories they represent. While the CS score can give us a basic understanding of the generation quality, it does not necessarily reflects the creative abilities of a model. For instance, if a model generates only canonical and mundane sketches of birds, then the generated sketches would more likely to be correctly classified by the trained Inception model.  In that case, the CS score will still be high. In contrast, the SDS score is more reliable in measuring the diversity of the generated sketch images.  
Tab.~\ref{tab:main_table} shows that DoodleFormer performs favourably against existing methods, in terms of SDS score, on both datasets. Fig.~\ref{fig:qual}(a)  shows a visual comparison of DoodleFormer with DoodlerGAN for creative sketch generation\textsuperscript{\ref{note1}}.

\subsection{User Study}
\begin{wrapfigure}{r}{0.5\textwidth} 
\vspace{-50pt}
  \begin{center}
    \includegraphics[width=0.5\textwidth]{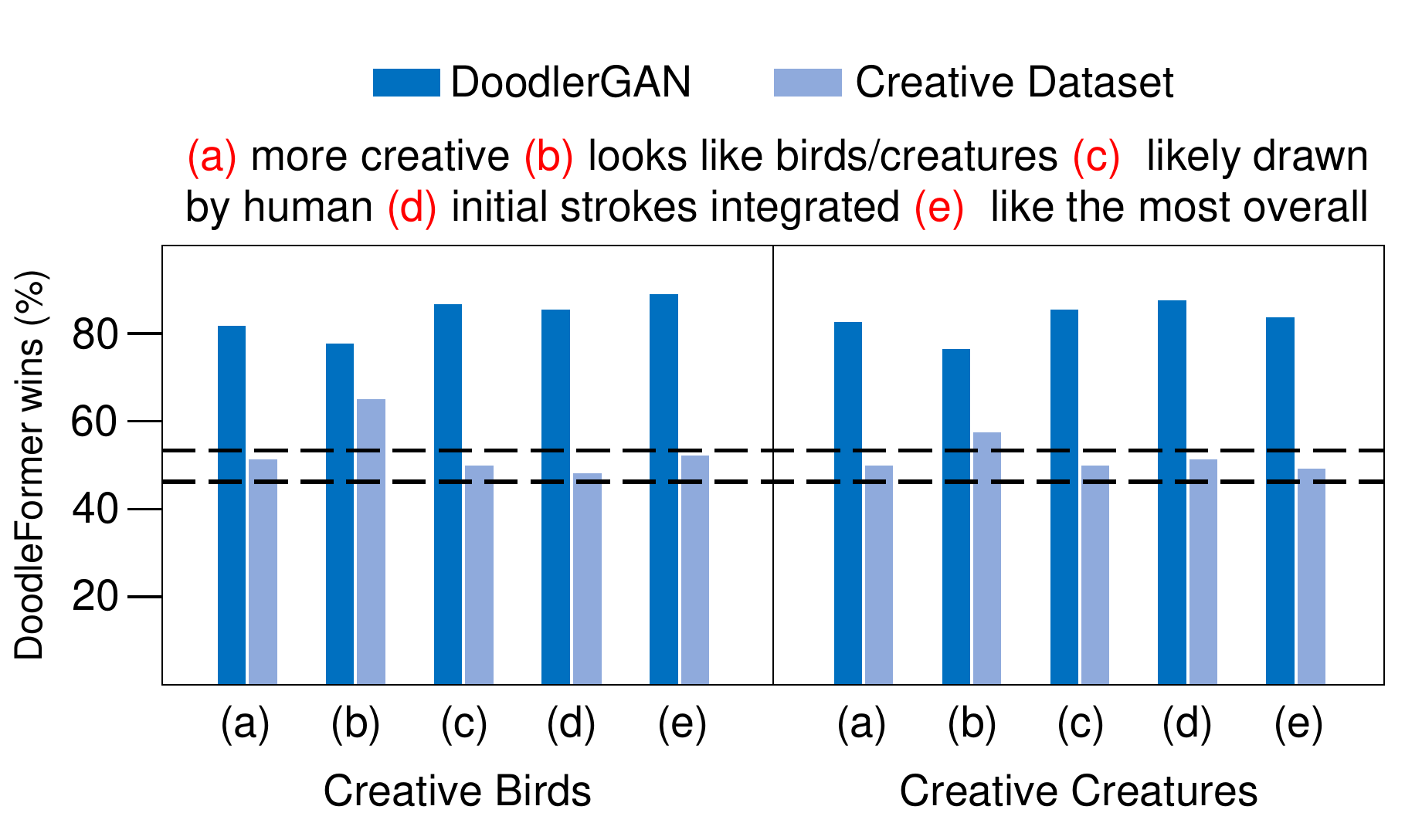}
    \caption{User study results on Creative Birds (left) and Creative Creatures (right) based on the five questions (\textcolor{red}{a-e}) mentioned in the legend. Higher values indicate  DoodleFormer is preferred more often over the compared approaches (DoodlerGAN and human drawn Creative datasets). 
    } 
    
    
    \label{fig:userstudy}
  \end{center}
  \vspace{-25pt}
  \vspace{1pt}
\end{wrapfigure}

Here, we present our user study to  evaluate the human plausibility of creative sketches generated by our DoodleFormer. 
Specifically, we show 100 participants pairs of sketches – one generated by DoodleFormer and the other by a competing approach. For each pair of images, similar to  DoodlerGAN  \cite{ge2020creative}, each participant is provided with 5 questions which are shown in the legend of Fig.~\ref{fig:userstudy} (a-e). DoodleFormer performs favorably against DoodlerGAN for all five questions on both datasets. For instance, DoodleFormer sketches were interpreted to be drawn by a human 86\%, having better initial strokes integration 85\% and being more creative 82\%, over DoodlerGAN on Creative Birds dataset. Further, for all the five questions, the DoodleFormer generated sketch images were found to be comparable with the human drawn sketches in Creative Datasets\footnote{\label{note1}Additional details and results are provided in supplementary material.}.

\subsection{Ablation Study}
We perform multiple ablation studies to validate the impact of proposed contributions in our framework. Tab.~\ref{tab:ablation1} shows the impact of two-stage framework, GAT blocks and GMM-based modeling on Creative Birds. Our single-stage baseline (referred as baseline$^{*}$) is a standard transformer-based encoder-decoder architecture, where initial strokes are given as input to the transformer encoder. The decoder sequentially generates all body parts which are then integrated to obtain final output sketch. The generated sketches using baseline$^{*}$ are unrealistic and suffer from body parts misplacement. The introduction of two-stage framework leads to an absolute gain of 26.3 in terms of FID score, highlighting the importance of a coarse-to-fine framework for realistic creative sketch generation. Our two-stage framework baseline neither uses the GAT block in encoders nor employs the GMM-based modeling in decoder. Instead of GMM-based modeling, we use a deterministic L1 loss in first stage (PL-Net) of two-stage baseline. While this two-stage baseline improves realism of generated sketches, it stills suffers from topological artifacts. The introduction of GAT block in encoders of two-stage baseline improves realism of generated sketches by capturing the structural relationship between different parts. Although GAT blocks improve FID score by a margin of 3.25, the generated sketches still lack diversity as indicated by only a marginal change in GD score. The introduction of GMM-based modelling in decoder improves diversity (\eg, appearance, size, orientation and posture) of generated sketches that leads to an absolute gain of 1.17 in GD (see also Fig.~\ref{fig:onecol}). Our final DoodleFormer (two-stage baseline + GAT blocks + GMM) achieves absolute gains of $30.0$ and $3.7$ in terms of FID score over  baseline$^{*}$ and two-stage baseline, respectively.

We also evaluate the design choices of our GAT blocks (see Tab.~\ref{tab:ablation2}). First, we replace GAT blocks in encoders with simple GCN layers. The standard transformer-based encoders outperforms this GCN-based baseline by a margin of 7.31. Further, we adapt the Mesh Graphormer~\cite{lin2021mesh} by using their Graphormer in encoders of our framework. 
Mesh Graphormer stacks transformer encoder layer and GCN block together in series. In our experiments, we observe this design based on loosely connected components performs slightly worse than standard transformer-based baseline. In contrast, an integrated design like ours performs comparatively better.

\begin{table}[t!]

\parbox{.47\linewidth}{
\begin{center}
\caption{\textbf{Impact of our two-stage framework, GAT blocks and GMM-based probabilistic modelling} on Creative Birds. 
}

\label{tab:ablation1}
\setlength{\tabcolsep}{3pt}
\scalebox{0.75}{
\begin{tabular}{l|l|c|c}
\toprule[0.4mm]

\rowcolor{mygray}
Design Choices & Methods & FID($\downarrow$) & GD($\uparrow$) \\ \midrule


Single-stage         & baseline$^{*}$ &46.45 &16.87\\\midrule

\multirow{3}{*}{Two-stage }       & baseline  &20.14  &17.05\\
        & baseline + GAT  &16.89 &17.16\\
      & baseline + GAT + GMM   &\textbf{16.45}  &\textbf{18.33}\\\bottomrule[0.4mm]
      
\end{tabular}
}
\end{center} \vspace{-0.8cm}

}
\hfil
\parbox{.47\linewidth}{
\begin{center}
\caption{\textbf{Comparison of alternative  design choices for the proposed GAT blocks} on Creative Birds. 
}
\label{tab:ablation2}
\setlength{\tabcolsep}{10pt}
\scalebox{0.75}{
\begin{tabular}{l|c|c}
\toprule[0.4mm]

\rowcolor{mygray}
Methods & FID($\downarrow$) & GD($\uparrow$) \\ \midrule

GCN layers        &27.45 &17.23\\
Transformer layers        &20.14 &17.05\\
Mesh Graphormer \cite{lin2021mesh}       &20.34 &16.78\\
\midrule
\textbf{GAT layers (ours) }      &\textbf{16.45}  &\textbf{18.33} \\  \bottomrule[0.4mm]
\end{tabular}
}
\end{center} \vspace{-0.8cm}
}
\end{table}

\subsection{Related Applications}

We also analyze DoodleFormer on three related tasks: 
user provided text to creative sketch generation, creative sketch completion and house layout generation. \\
\noindent\textbf{Text to Creative Sketch Generation: } Here, the text description is given as conditional input to encoder $E_{c}$ in PL-Net, yielding a coarse structure of desired sketch which is fed to PS-Net to generate final sketch. 
We remove $\mathcal{R}_E$ from the PS-Net, and the $cls$ token output from the encoder $\Bar{E}_c$ is directly passed as input to $\mathcal{R}_D$.
We use 80\%-20\% train-test split.
We compare DoodleFormer with two popular text-to-image methods: StackGAN~\cite{zhang2017stackgan}, AttnGAN~\cite{xu2018attngan} on Creative Birds and Creative Creatures. DoodleFormer performs favorably against these methods in terms of FID and GD scores on both datasets. On Creative Birds, StackGAN, AttnGAN and DoodleFormer achieve respective FID scores of [53.1, 45.2, \textbf{18.5}], and GD scores of [16.7, 16.5, \textbf{17.3}].
Fig.~\ref{fig:qual} (c) shows a qualitative  comparison\textsuperscript{\ref{note1}}.



\noindent\textbf{Creative Sketch Completion: } 
Given an incomplete sketch as input, DoodleFormer attempts to creatively complete the rest of the sketch. First, PL-Net obtains the bounding boxes for missing parts. Then, PS-Net generates an image containing the required missing parts which is then integrated with the incomplete sketch input to obtain the final output. On both Creative Birds and Creative Creatures, DoodleFormer achieves favorable results compared to DoodlerGAN in terms of FID and GD scores. 
On Creative Birds, DoodlerGAN and DoodleFormer achieve respective FID scores of [44.2 and \textbf{18.3}] and GD scores of [15.1 and \textbf{17.8}].
Fig.~\ref{fig:qual} (b) shows the qualitative comparison\textsuperscript{\ref{note1}}.



\begin{table}[t!]
\begin{center}
\caption{\textbf{House Layout Generation:} We compare our approach with the existing methods in terms of FID and Compatibility
scores (obtained by the graph edit distance). The dataset samples are split into five groups based on the room counts (1-3, 4-6, 7-9, 10-12, and 13+).
}\vspace{0.2cm}
\label{tab:house}
\setlength{\tabcolsep}{6pt}
\scalebox{0.75}{
\begin{tabular}{l|c|c|c|c|c|c|c|c|c|c}
\toprule[0.4mm]

\rowcolor{mygray} 
\cellcolor{mygray} &
  \multicolumn{5}{c|}{\cellcolor{mygray}FID ($\downarrow$)} &
  \multicolumn{5}{c}{\cellcolor{mygray}Compatibility ($\downarrow$)} \\ 
\rowcolor{mygray} 
\multirow{-2}{*}{\cellcolor{mygray}Methods} &
  1-3 & 4-6 & 7-9 & 10-12 & 13+ & 1-3 & 4-6 & 7-9 & 10-12 & 13+\\ \midrule
  
Ashual \etal~\cite{ha2017neural}       &64.0 &92.2 &87.6 &122.8 &149.9 &0.2 &2.7 &6.2 &19.2 &36.0 \\
Johnson \etal~\cite{Karras2019stylegan2}     &69.8 &86.9 &80.1 &117.5 &123.2 &0.2 &2.6 &5.2 &17.5 &29.3 \\
House-GAN~\cite{ge2020creative}      &13.6 &\textbf{9.4} &14.4 &11.6 &20.1 &0.1 &1.1 &2.9 &3.9 &10.8 \\\midrule
\textbf{Ours}       &\textbf{9.6} &10.1 &\textbf{11.2} &\textbf{9.7} &\textbf{18.2} &\textbf{0.1} &\textbf{1.0} &\textbf{2.1} &\textbf{2.4} &\textbf{8.3}    \\  \bottomrule[0.4mm]
\end{tabular}

}
\end{center} \vspace{-0.5cm}
\end{table}

\noindent\textbf{House Layout Generation: }  Finally, we use our proposed PL-Net architecture for the house-plan generation \cite{nauata2020house} task. The goal is to take a bubble diagram as an input, and generate a diverse set of realistic and compatible house layouts. A bubble diagram is represented by a graph where each node contains information about rooms and edges indicate their spatial adjacency.  The output house layout is represented as axis-aligned bounding boxes. The Encoder $E_c$ takes the room type information as input and Probabilistic Decoder subsequently outputs boundary boxes for each room. To transform the obtained boundary box layout to a floor plan layout, we employ a floor-plan post-processing strategy. In this process, we first extract boundary lines of the generated boundary boxes. Next, we merge the adjacent line segments together and further align them to obtain a closed polygon. 

\begin{figure*}[t!]
\begin{center}
   \includegraphics[width=\linewidth]{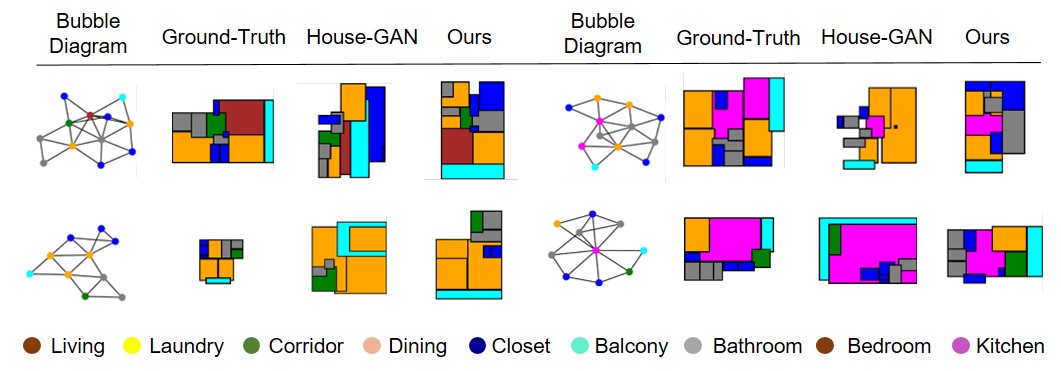}\vspace{-0.5cm}
\end{center}
\caption{Qualitative results of House Layout Generation. Given the input bubble diagram, We compare the house layout sample generated using our method with the House-GAN~\cite{nauata2020house}. Our method produces house layouts that are well
aligned with input bubble diagram texts. 
}
\vspace{-0.4cm}
\label{fig:hsgan}
\end{figure*}

We perform the house-plan generation experiments on LIFULL HOME's dataset \cite{Lifull}. For fair comparison, we follow the same setting used by House-GAN~\cite{nauata2020house}. We divide the samples into five groups based on the number of rooms: 1-3, 4-6, 7-9, 10-12, and 13+. To test the generalization ability in each group, we train a model while excluding samples in the same group. At test time, we randomly pick a bubble input diagram from each group and generate $10$ samples. Similar to House-GAN~\cite{nauata2020house}, we quantitatively measure the performance of our method in terms of FID and compatibility scores. The compatibility score is the graph editing distance \cite{abu2015exact} between the input bubble diagram and the bubble diagram constructed from the output layout. Tab.~\ref{tab:house} shows that our method outperforms existing house-plan generation methods both in terms of FID and compatibility scores. Fig.~\ref{fig:hsgan} shows the qualitative comparison of our house layout generation approach with House-GAN~\cite{nauata2020house}. 

\section{Conclusion}

We proposed a novel coarse-to-fine two-stage approach, DoodleFormer, for creative sketch generation. We introduce graph-aware transformer encoders that effectively capture global dynamic as well as local static structural relations among different body parts. To ensure diversity of generated creative sketches, we introduce a probabilistic coarse sketch decoder that explicitly models variations of each sketch body part to be drawn. We show the effectiveness of DoodleFormer on two datasets by performing extensive qualitative, quantitative and human-based evaluations. In addition, we demonstrate promising results on related applications such as text to creative sketch generation, sketch completion and house layout generation.

\pagestyle{headings}
\mainmatter
\def\ECCVSubNumber{6722}  
\title{DoodleFormer: Creative Sketch Drawing with Transformers\\Supplementary Material}

\titlerunning{ECCV-22 submission ID \ECCVSubNumber} 
\authorrunning{ECCV-22 submission ID \ECCVSubNumber} 
\author{Anonymous ECCV submission}
\institute{Paper ID \ECCVSubNumber}


\titlerunning{DoodleFormer: Creative Sketch Drawing with Transformers\\Supplementary Material}
%
\authorrunning{A. K. Bhunia \etal}

\author{Ankan Kumar Bhunia$^1$, Salman Khan$^{1,2}$, Hisham Cholakkal$^{1}$,
Rao Muhammad Anwer$^{1,3}$, Fahad Shahbaz Khan$^{1,4}$, Jorma Laaksonen$^{3}$, Michael Felsberg$^{4}$}
\institute{
  $^1$~Mohamed bin Zayed University of AI, UAE \quad
  $^2$~Australian National University, Australia
  $^3$~Aalto University, Finland
  $^4$~Link{\"o}ping University, Sweden
\texttt{\small ankan.bhunia@mbzuai.ac.ae} 
  }


\maketitle
 
   \begin{center}
    \includegraphics[width=1\linewidth]{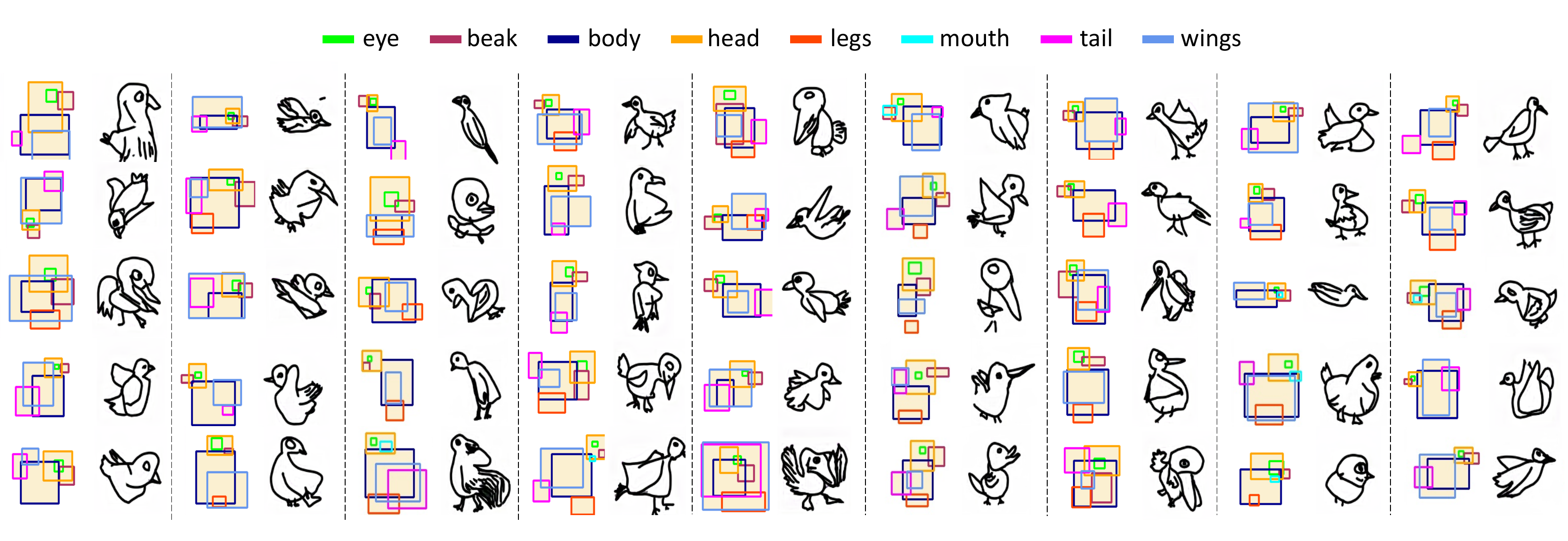}
    \captionof{figure}{Visualization of the \emph{coarse-to-fine} creative sketch drawing process. The model first draws the holistic coarse structure of the sketch and then
 fills the fine-details to generate the final sketch.  Both the coarse structure and the final sketch output are shown side-by-side in the figure. By first drawing the holistic coarse structure of the sketch aids to appropriately decide the location and the size of each sketch body part to be drawn. }\label{fig:bbox}
  \end{center}



In this supplementary material, we present additional qualitative results and additional user study details. In Sec.~\ref{sec:coarse}, we present the visualizations depicting the coarse-to-fine sketch generation process . Sec.~\ref{sec:metrics} presents the additional details for the calculation of the evaluation metrics. Sec.~\ref{sec:qual_add} shows additional quantitative results.
Sec.~\ref{sec:user} provides additional details of user study experiments. 

 \section{Coarse-to-fine Sketch Generation}
  \label{sec:coarse}
Fig.~\ref{fig:bbox} presents example visualizations depicting the intermediate results of our \emph{coarse-to-fine} creative sketch generation. The proposed two-stage DoodleFormer framework decomposes the creative sketch generation
problem to first capture the holistic coarse structure of the sketch and then injecting fine-details to generate the final sketch. Both the coarse structure from the first-stage PL-Net, and the final sketch output from the second-stage PS-Net, are shown side-by-side in Fig.~\ref{fig:bbox}. By first drawing the holistic coarse structure of the sketch aids to appropriately determine the location and the size of each sketch body part to be drawn. 

 \begin{figure*}[t!]
\begin{center}
   \includegraphics[width=1\textwidth]{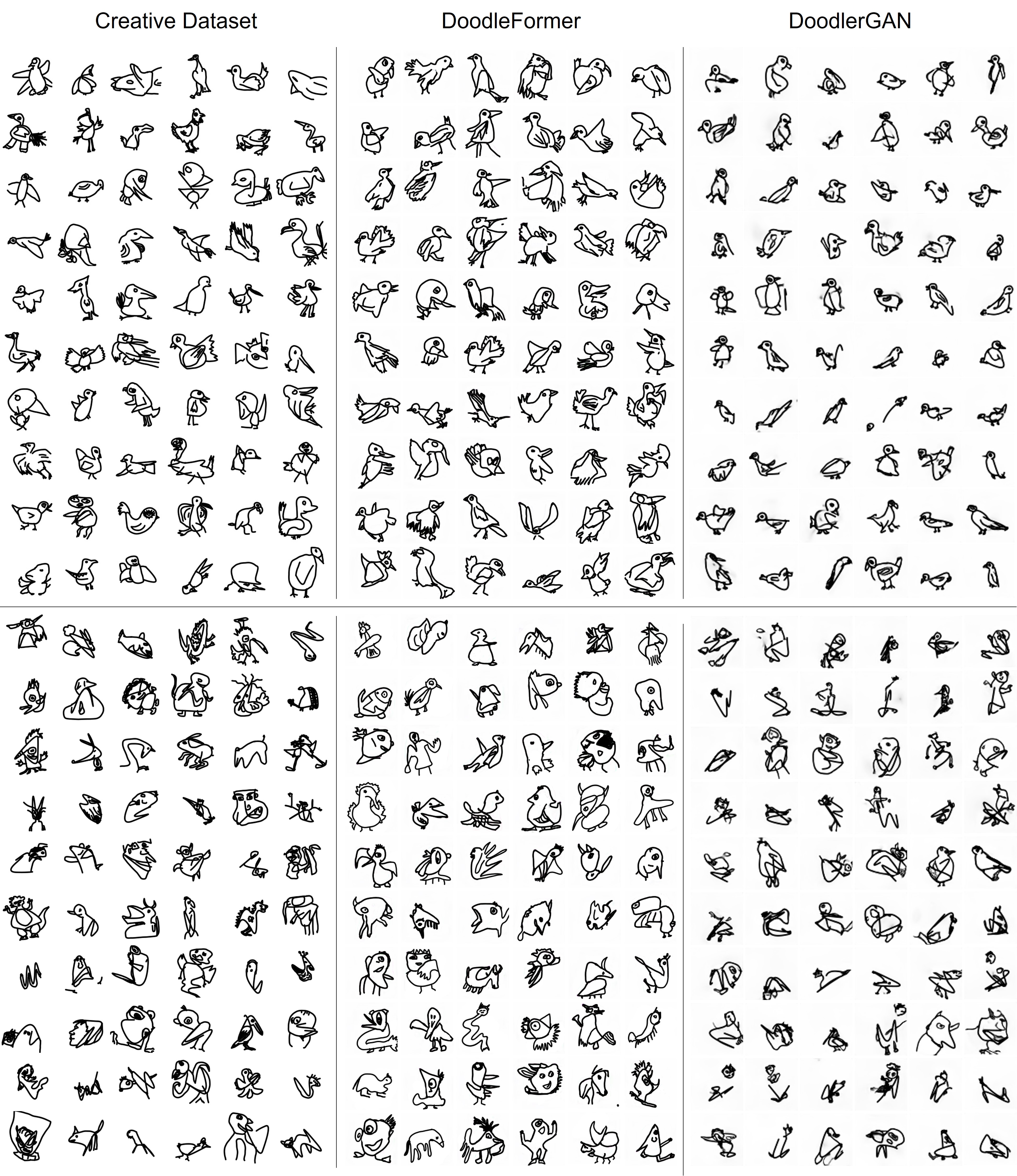}\vspace{-0.5cm}
\end{center}
\caption{Additional qualitative comparisons on Creative Birds and Creative Creatures datasets. In this figure, we compare the generated sketches using the proposed DoodleFormer (in the middle column) with DoodlerGAN \cite{ge2020creative} (in the right column). The human drawn creative sketch images from the datasets are shown in the left column. DoodlerGAN suffers from topological artifacts. Also, DoodlerGAN generated sketches have lesser diversity in
terms of size, appearance and posture. The proposed DoodleFormer alleviates the issues of topological artifacts and disconnected body parts, generating creative sketches that are more realistic and diverse.
} \vspace{-0.4cm}
\label{fig:qual360}
\end{figure*}

 \section{Additional Details of Evaluation Metrics}
 \label{sec:metrics}
 We quantitatively evaluate our proposed approach based on four metrics:  Fr\`echet inception distance (FID), generation diversity (GD), characteristic score (CS), and semantic diversity score (SDS). Following~\cite{ge2020creative}, we use an inception model trained on the QuickDraw3.8M dataset ~\cite{xu2020deep} to calculate these metrics. We generate $10,000$ sketches based on a randomly sampled set of previously unseen initial strokes. Then, the generated sketches are resized into $64 \times 64$ images and passed through the trained inception model to calculate the above mentioned metrics.
 
\begin{figure*}[t!]
\begin{center}
   \includegraphics[width=1\textwidth]{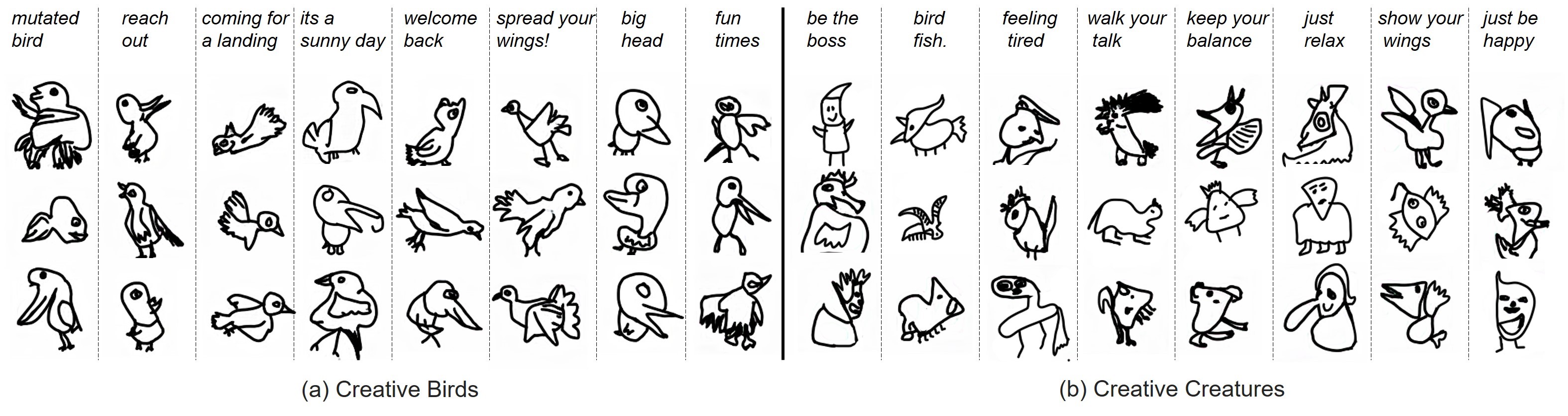}\vspace{-0.5cm}
\end{center}
\caption{Additional qualitative results for text to creative sketch generation of DoodleFormer on Creative Birds and Creative Creatures dataset. For each input text, we show three different samples generated using DoodleFormer. The results demonstrate that our approach is capable of generating diverse sketch images while ensuring that the generated sketches are well matched with the user provided text inputs. 
} \vspace{-0.4cm}
\label{fig:text}
\end{figure*}

\begin{figure*}[t!]
\begin{center}
   \includegraphics[width=1\textwidth]{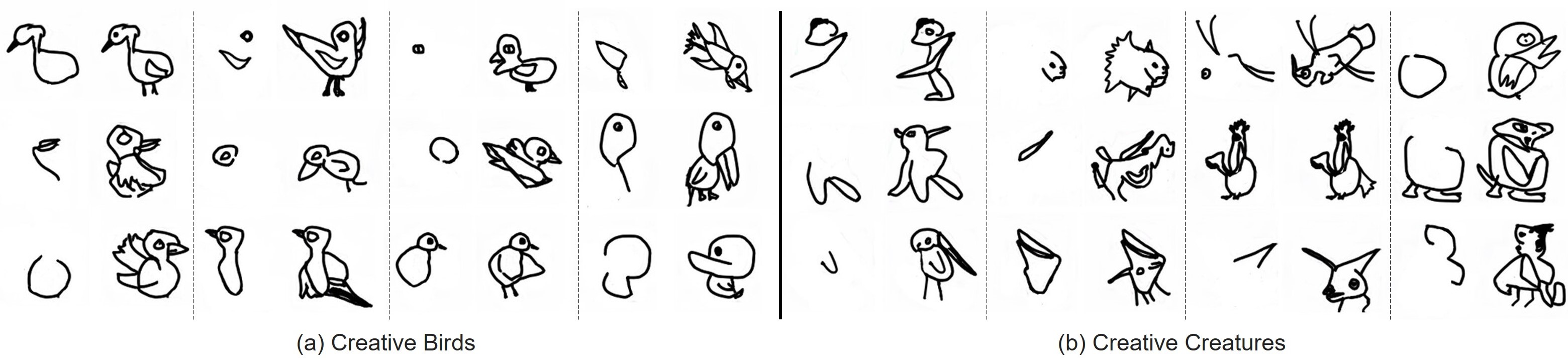}\vspace{-0.5cm}
\end{center}
\caption{Additional qualitative results for text to creative sketch completion of DoodleFormer on Creative Birds and Creative Creatures dataset. The results depict that DoodleFormer accurately completes the missing regions of the given partial input sketch.
} \vspace{-0.4cm}
\label{fig:complete}
\end{figure*}
 
 \section{Additional Qualitative Results}
 \label{sec:qual_add}
 Fig.~\ref{fig:qual360} shows additional qualitative results of our proposed DoodleFormer for creative sketch generation on both datasets (Creative Birds and Creative Creatures). For better visual comparison, we also show the creative  sketch images from the datasets and the DoodlerGAN \cite{ge2020creative} generated sketch images in the same figure. 
 
\begin{figure}[t!]
\begin{center}
   \includegraphics[width=0.7\linewidth]{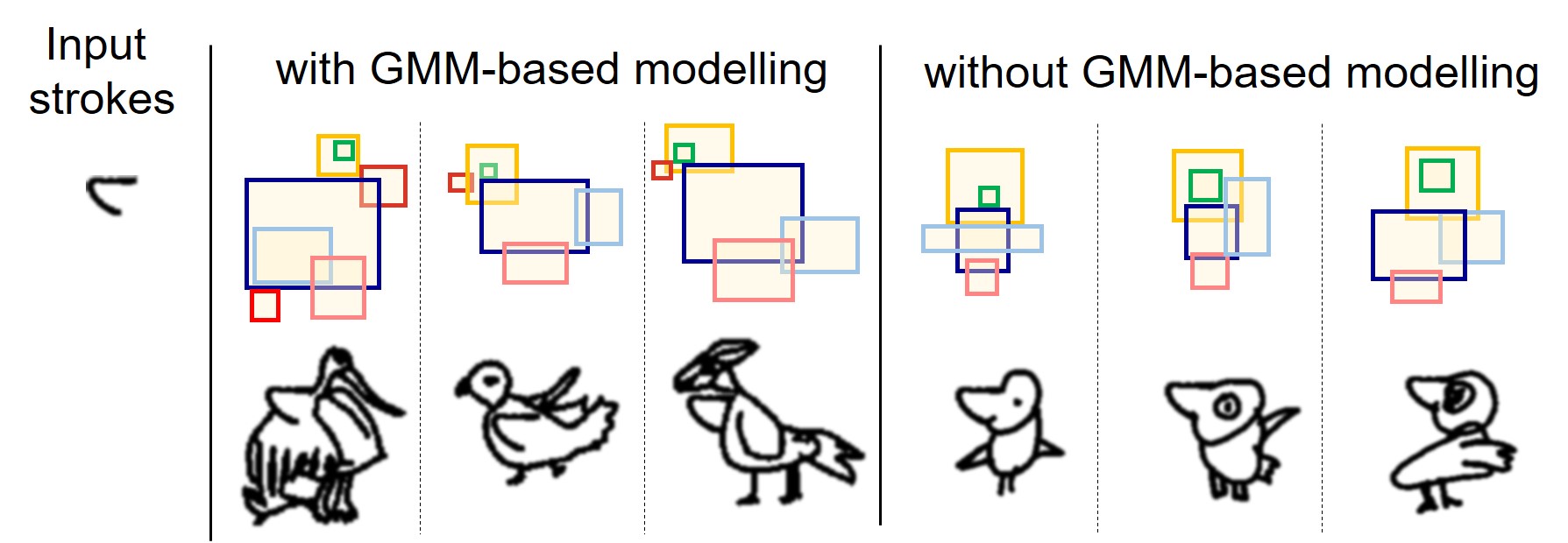}
\end{center}
\vspace{-0.45cm}
\caption{Qualitative analysis of GMM-based modelling for common initial strokes. The introduction of the GMM-based modelling substantially improves the diversity of generated sketches. }\vspace{-0.45cm}
\label{fig:gmm_extra}
\end{figure}

\begin{figure}[t!]
\begin{center}
   \includegraphics[width=0.7\linewidth]{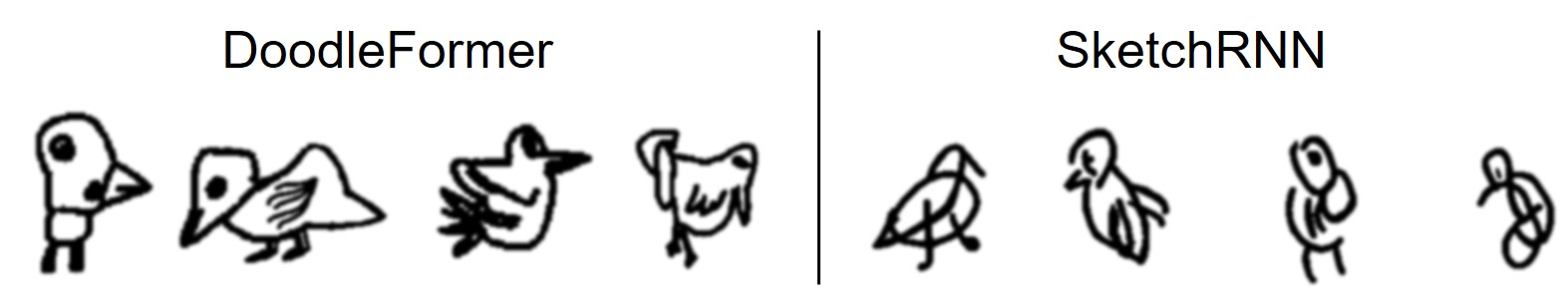}
\end{center}
\vspace{-0.55cm}
\caption{A visual comparison of creative sketches generated by DoodleFormer and SketchRNN~\cite{ha2017neural} on Creative Birds dataset. }\vspace{-0.45cm}
\label{fig:sketchrnn}
\end{figure}

 In Fig.~\ref{fig:text}, we present additional qualitative results for text to creative sketch generation on both datasets (Creative Birds and Creative Creatures). For each user provided input text, we show three samples generated using our proposed DoodleFormer. The results demonstrate the effectiveness of our approach towards generating diverse sketch images while ensuring that the generated sketches are well matched with the user provided text inputs. Fig.~\ref{fig:complete} shows qualitative results for creative sketch completion.  The results show that DoodleFormer accurately completes the missing regions of these challenging incomplete sketches. Fig.~\ref{fig:gmm_extra} shows
that GMM-based modelling can generate multiple bound-
ing box layouts with more diverse sketch results (w.r.t appearance, size, orientation and posture).  Fig.~\ref{fig:sketchrnn} presents a qualitative comparison of DoodleFormer with SketchRNN~\cite{ha2017neural}.
 


 \section{User Study Additional Details}
\label{sec:user}
 Here, we present additional details of our user studies on 100 human participants to evaluate the creative abilities of our proposed DoodleFormer. We compare the DoodleFormer generated sketches with DoodlerGAN and human-drawn Creative dataset sketches. Specifically, we show participants pairs of sketches – one generated by DoodleFormer and the other by a competing approach. Each participant has to answer 5 questions for each of these pairs. The questions are: which one (a) is more creative? (b) looks more like a bird/creature? (c) is more likely to be drawn by a human? (d) in which case, the initial strokes are well integrated? (e) like the most overall. 
 Our proposed DoodleFormer performs favorably against DoodlerGAN \cite{ge2020creative} for all five questions on both datasets. Further, the DoodleFormer generated sketch images were observed to be comparable with the sketches drawn by human in Creative Datasets for all the five questions. 
\\
\textbf{Acknowledgements.} This work has been funded by the Academy of Finland in project USSEE (345791) and supported by the Aalto Science-IT project.

\bibliographystyle{splncs04}
\bibliography{egbib}
\end{document}